\UseRawInputEncoding
\documentclass[review,12pt,a4papper]{elsarticle}
\usepackage{geometry}
\usepackage{amsmath,amssymb}

\usepackage{multirow}
\usepackage{float}
\usepackage{makecell}
\usepackage{algorithm}
\usepackage{algorithmic}
\usepackage{graphicx}
\usepackage{bm}
\usepackage{subfigure}
\usepackage{booktabs}
\usepackage{array, caption, threeparttable}
\usepackage{appendix}
\usepackage[section]{placeins}
\usepackage[font=small,labelfont=bf]{caption}
\usepackage{lineno,hyperref}
\modulolinenumbers[5]

\begin{document}

\begin{frontmatter}

\title{Multi-view Feature Extraction based on Triple Contrastive Heads}

\author{Hongjie Zhang}
\address{College of Information and Electrical Engineering, China Agricultural University, Beijing 100083, China}

\begin{abstract}
	Multi-view feature extraction is an efficient approach for alleviating the issue of dimensionality in highdimensional
	multi-view data. Contrastive learning (CL), which is a popular self-supervised learning method, has
	recently attracted considerable attention. 
    In this study, we propose a novel multi-view feature extraction method based on triple contrastive heads, which combines the sample-, recovery- , and  feature-level contrastive losses to extract the sufficient yet minimal subspace discriminative information in compliance with information bottleneck principle. In MFETCH,
	we construct the feature-level contrastive loss, which removes the redundent information in the consistency information to achieve the minimality of the subspace discriminative information. Moreover, the recovery-level contrastive loss is also constructed in MFETCH, which captures the view-specific discriminative information to achieve the sufficiency of the subspace discriminative information.The numerical experiments demonstrate that the proposed method offers a strong
	advantage for multi-view feature extraction.\par 
\end{abstract}


\end{frontmatter}


\section{Introduction}
Multi-view data contains richer feature information than single-view data, as it is described from distinct aspects. Therefore, the effectiveness of classification and clustering can be improved by using multi-view data\cite{c1,cluster1,cluster2,e1,s1}. However, multi-view data is often high-dimensional, which leads to some drawbacks for training machine learning tasks. For example, high-dimensional multi-view data results in a considerable waste of time and costs as well as causes the problem known as the “curse of dimensionality”.  Currently, multi-view feature extraction is an effective method for solving high-dimensional problem\cite{ex1,ex2,ex3}, which transforms the original samples into a low-dimensional subspace by some projection matrices. Although the effect of feature extraction is often slightly worse than that of deep
learning, it has always been a research hotspot owing to its
strong interpretability and compatibility with any type of hardware
(CPU, GPU, and DSP). Therefore, there is an urgent need for developing
traditional multi-view feature extraction methods to extract discriminative
features effectively. \par 
In the field of deep learning, contrastive learning (CL)\cite{cl1,cl2,cl3,cl4,cl5} has been widely applied to process multi-view data, and a large number of existing CL-based methods have shown excellent performance in unsupervised learning. The CL-based methods extract subspace discriminative information (useful for downstream sample discrimination) by promoting the consistency of any two cross views. Specifically, InfoNCE loss based on CL is proposed in contrastive predictive coding\cite{cpc}, which obtains the consistency information by maximizing the similarity of positive pairs and minimizing the similarity of negative pairs in the embedding space.
Subsequently, a large number of CL-based studies have focused on the methods of constructing positive and negative pairs, so as to extract the consistency information suitable for subspace discrimination. Tian et al. proposed contrastive multiview coding (CMC)\cite{cmc}, which constructs the same sample in any two views as positive pairs and distinct samples as negative pairs, and subsequently, optimizes a neural network by minimizing the InfoNCE loss. Chen et al. proposed a simple framework for CL (SimCLR)\cite{simclr}, which constructs positive and negative pairs similar to CMC by performing data augmentation for single-view data. Due to the performance of CL is affected by the number of negative pairs, a number of CL-based methods were proposed to increase the batch size or memory banks. For example, non-parametric softmax classifier\cite{npid} and pretext-invariant representation learning\cite{pirl} use memories that contain the whole training set, while the momentum contrast\cite{moco} keeps a queue with features of the last few batches as memory. However, increasing the number of negative pairs results in a waste of time and storage, and it has been shown that hard negative pairs improve the performance of CL better than increasing the number of negative pairs. Therefore, invariance propagation (InvP)\cite{invp} and mixing of contrastive hard negatives (MoCHi)\cite{mochi} based on distinct hard sampling strategies were proposed. InvP defines the positive and negative pairs with structural information, while MoCHi constructs the hardest negative pairs and even harder negative pairs in the original space.  In addition, nearest-neighbor contrastive learning of visual representation (NNCLR)\cite{nnclr} and consistent contrast (CO2)\cite{co2} utilize distinct strategies of top-k nearest neighbors to avoid pseudo-negative pairs when optimizing the network parameters.
Prototypical contrastive learning (PCL)\cite{pcl} and hiearchical contrastive selective coding (HCSC)\cite{hcsc} construct positive and negative pairs from the level of clustering. PCL introduces prototypes as latent variables to help find the maximum-likelihood estimation of the network parameters in an Expectation-Maximization framework, while HCSC propose to represent the hierarchical semantic structures of image representations by dynamically maintaining hierarchical prototypes.\par
\begin{figure*}[!t]
	\centering
	\subfigure[]{\begin{minipage}[c]{0.45\textwidth} \centering\includegraphics[width=.6\textwidth]{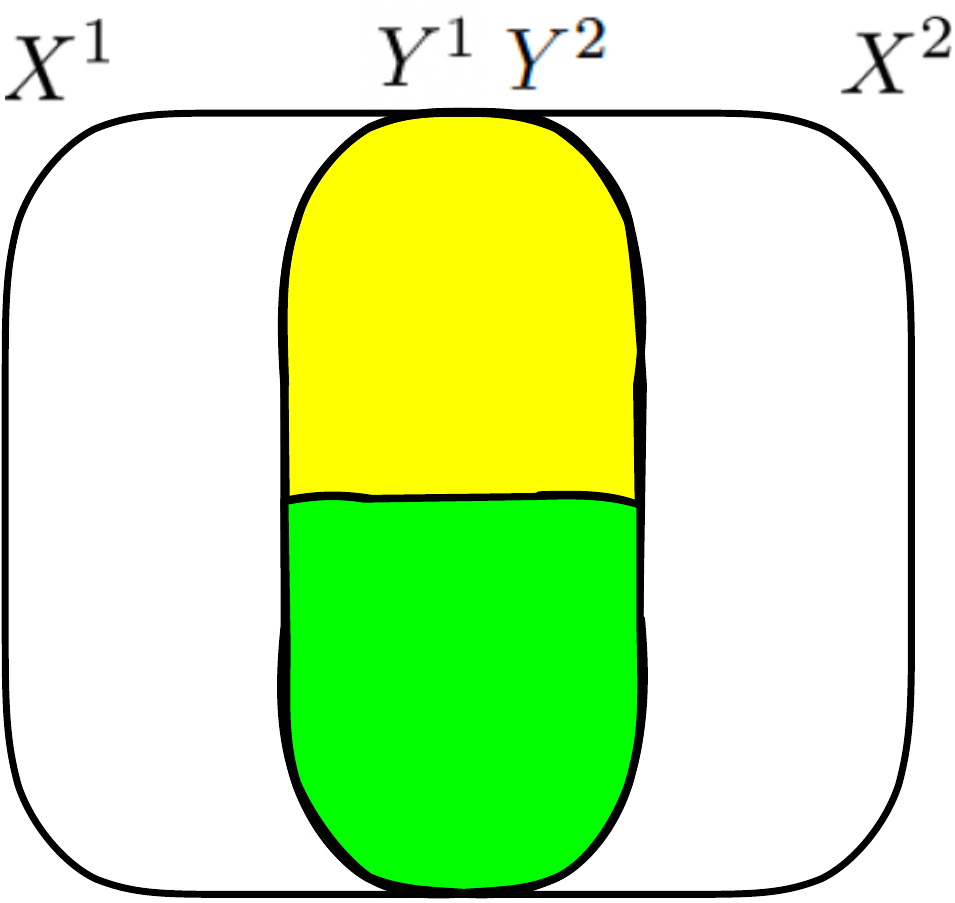}\end{minipage}}
	\subfigure[]{\begin{minipage}[c]{0.45\textwidth} \centering\includegraphics[width=0.6\textwidth]{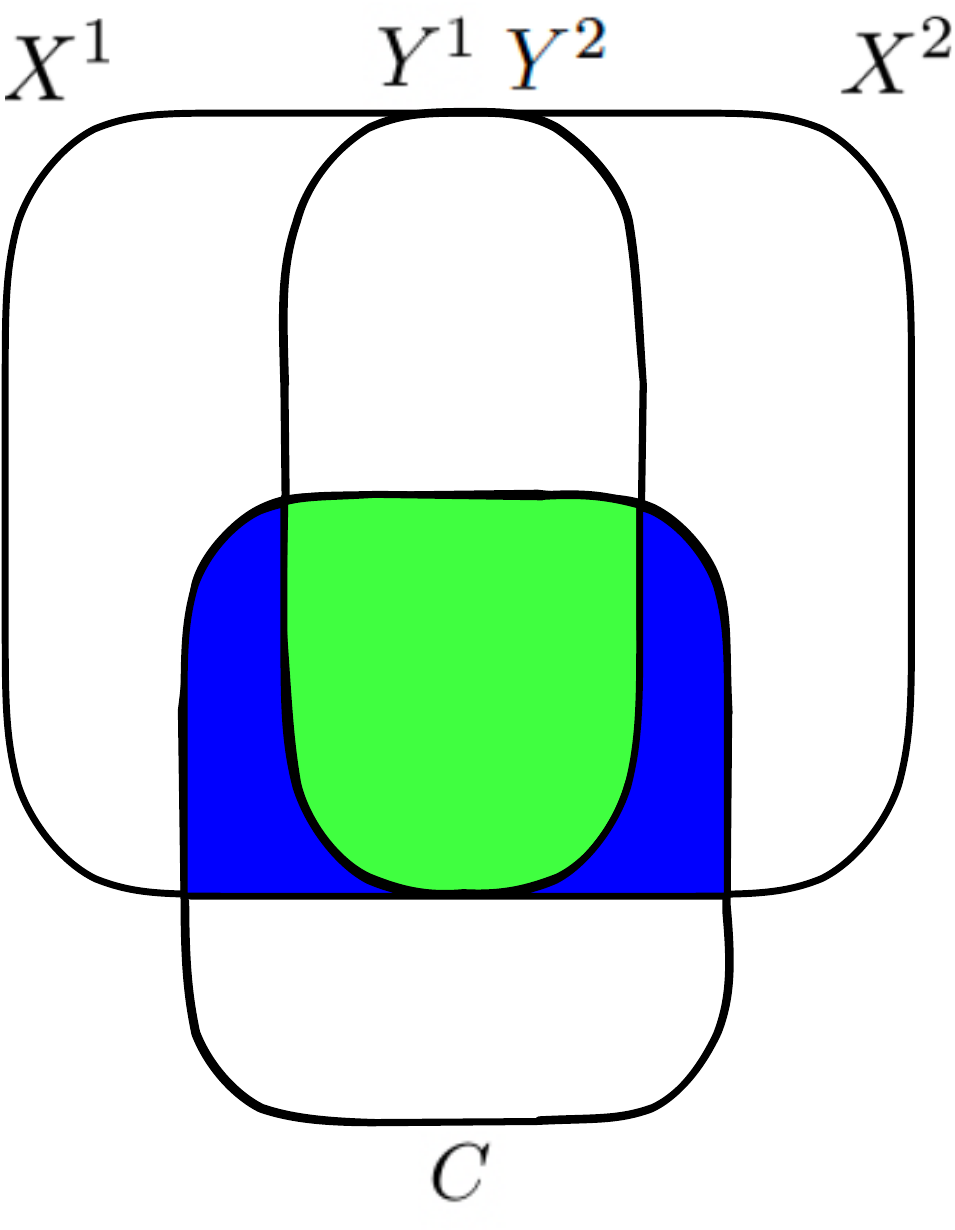}\end{minipage}}
	\caption{Illustration of the relation two views $X^1$ and $X^2$, label $C$ in terms of information entropy. (a) The consistency information is extracted by sample-level CL (yellow and green areas), which cantains discriminative information (green area) and redundant information (yellow area). (b) The sample-level CL does not capture the view-specific discriminative information (blue area).} 
	\label{fig1}
\end{figure*}

 
However, the above CL-based methods are used for optimizing the network parameters in the field of deep learning, they can not obtain the projection matrices to process the traditional multi-view feature extraction problem. In addition, the above methods promote the consistency of any two cross views from the perspective of the subspace samples (abbreviated as sample-level CL), which contains the redundant information (repeated representations of the same information in distinct features) suppressing the important discriminative information. More importantly, these methods do not capture the view-specific discriminative information.  A more intuitively illustration in terms of information entropy is provided in the Figure \ref{fig1}.\par 
Recently, the information bottleneck (IB) principle has been introduced into the field of deep learning, which encourages extracting sufficient yet minimal discriminative information for unsupervised learning. 
However, labels are not provided in unsupervised learning, so we cannot directly calculate the sufficient yet minimal subspace discriminative information.
In this study, we propose a novel multi-view feature extraction method based on triple contrastive heads (MFETCH), which combines the sample-,  feature- and recovery-level contrastive losses to extract the sufficient yet minimal subspace discriminative information in accordance with the IB principle. 
Specifically, a feature-level contrastive loss is constructed in our MFETCH, which removes the redundant information in the consistency information to help sample-level contrastive loss achieve the minimality of the subspace discriminative information. 
More importantly, a recovery-level contrastive loss is also constructed, which captures the view-specific discriminative information to achieve the sufficiency of the subspace discriminative information.
In addition, the parameter $\alpha$ and $\beta$ is introduced into MFETCH to balance the importance of sample-, recovery- and feature-level contrastive losses.\par 
The main contributions of this study are as follows:
\begin{itemize}
	\item A novel unsupervised multi-view feature extraction method based on triple contrastive heads is proposed, which combines the sample-,  feature- and recovery-level contrastive losses to extract the sufficient yet minimal subspace discriminative information in accordance with the IB principle.
	\item The feature-level contrastive loss is constructed to remove the redundant information, which achieves the minimality of the subspace discriminative information.
	\item The recovery-level contrastive loss is constructed to capture the view-specific discriminative information, which achieves the sufficiency of the subspace discriminative information.
	\item The experiments on four real-world datasets show the advantages of the proposed method.
\end{itemize}

The remainder of this article is organized as follows. Traditional multi-view  feature extraction methods are briefly introduced in Section 2. Subsequently, the method and analysis are discussed in Section 3. Extensive experiments that were conducted on several real-world datasets are outlined in Section 4. Finally, the conclusion of this study are presented in Section 5.

\section{Related Work}
In unsupervised multi-view feature extraction, one of the most classical methods is canonical correlation analysis (CCA)\cite{cca}. It obtains two projection matrices by maximizing the correlation of distinct representations of the same subspace sample in two views, which explores actually the consistency information. However, CCA is only suitable for processing linear data. By combining kernel trick with traditional CCA, kernel canonical correlation analysis was proposed to process nonliner two-view data. With the development, a priori manifold structure is considered beneficial for multi-view feature extraction, so local preserving canonical correlation analysis (LPCCA)\cite{lpcca} and a new LPCCA (ALPCCA)\cite{alpcca} were proposed to preserve local neighbor relations when obtaining the projection matrices. The difference is that the former finds projection matrices by preserving local neighbor information in each view, while the latter explores extra cross-view correlations between neighbors. However, the selection of the neighbors is determined by human experience, which may produce incorrect structural information. To explore a more realistic structural information, canonical sparse cross-view correlation analysis \cite{cscca} was proposed, which introduces sparse
reconstruction into LPCCA to obtain local intrinsic geometric
structure automatically. On the basis of this, Zhu et al. proposed a weight-based CSCCA \cite{wcscca} which measures the correlation between two
views using the cross-view weighting information. In addition,  Zhao et al. proposed co-training
locality preserving projections\cite{colpp} which aims at finding
the subspace samples by promoting the consistency of the local neighbor structures in two views.\par 
However, all of the above methods are only suitable for
two-view data. Multi-view CCA\cite{mcca} was proposed to process multi-view data by maximizing the total canonical correlation coefficients of any two cross views. Based on single-view laplacian eigenmaps\cite{le},  multi-view spectral embedding (MSE)\cite{mse} is proposed, which obtains the projection matrices in order to make the embedded samples are 
smooth. Wang et al. proposed a new method named multi-view reconstructive
preserving embedding (MRPE)\cite{mrpe} by using a novel locality linear embedding
scheme. The embedded samples are obtained by MSE, MRPE and CMSRE directly, but they can not obtain the embedding of a new
sample. Therefore, sparsity preserving multiple canonical correlation
analysis\cite{spmcca} and graph multi-view canonical
correlation analysis\cite{gmcca} were proposed to solve the problem. As a further extension, Cao et al. proposed multi-view partial least squares\cite{mvpls}. In addition, Wang et al. proposed kernelized multi-view subspace analysis\cite{kmsa} which can automatically learn the weights for all views and extract the nonlinear features simultaneously.\par
Although the above methods explore the consistency information by various ways, this still contains the redundant information, which can suppress the important discriminative features. More importantly, the above methods do not capture the view-specific discriminative information, which result in failure to provide sufficient discriminative information for downstream tasks. Therefore, neither the traditional CL-based methods nor the traditional multi-view feature extraction methods can extract the sufficient yet minimal subspace discriminative information in accordance with the IB principle.

\section{Method and Analysis}
\subsection{Notation and Definition}
This study considers the problem of unsupervised multi-view  feature extraction. Let us mathematically formulate this problem as follows. \par 
Multi-view feature extraction problem: Given $V$ training sample sets $X^m=[x_1^m,x_2^m,...,x_n^m]\in R^{D_m\times n}, m=1,2,...,V$ from $V$ different views, where $X^m , m=1,2,...,V$ represents the dataset from the $m$th view, $x_i^m, i=1,2,...,n$ represents the $i$th sample in the $m$th view. $D_m$ represents the feature dimension in the $m$th view. The purpose of feature extraction is to find projection matrices $P_m\in R^{D_m\times d}, m=1,2,...,V$ to derive the low-dimensional embeddings $Y^m=[y_1^m,y_2^m,...,y_n^m]\in R^{d\times n}$ for $X^m$ calculated by $P_m^TX^m$, where $d\ll D^m$.\par
\subsection{Method}
\subsection*{\rm{\textbf{Sample-level CL}}}
The basic idea of sample-level CL is to optimize the neural network by exploring the consistency information from the perspective of the subspace samples, which maximize the similarity of positive pairs ($x_i^m, x_{i}^{m +}$) and minimize the simility of negative pairs ($x_i^m, x_{i}^{m -}$) in the InfoNCE loss. The main difference between current CL-based methods is the strategies for constructing positive and negative pairs.  To perform traditional multi-view feature extraction, the InfoNCE loss is specified as follows:
\begin{equation}
\begin{aligned}\label{eq1}
\mathcal L^\theta_{\text {InfoNCE}}(P_m,m=1,...,V)=
\sum_{m=1}^{V}\mathbb{E}_{x^m_i, x_i^{m +}, x_{i}^{m-}}\left[-\log \frac{\sum_{x_i^{m+}}\operatorname{exp}(\operatorname{sim}\left(y^m_i,  y_i^{m +}\right)}{\sum_{x_i^{m+}}\operatorname{exp}(\operatorname{sim}\left(y_i^m,  y_i^{m +}\right)+ \sum_{x_i^{m-}}\operatorname{exp}(\operatorname{sim}\left(y_i^m,  y_i^{m -}\right)}\right],
\end{aligned}
\end{equation}
where 
\begin{equation}
\operatorname{sim}\left(y_i^m, y_i^{m +}\right)=\frac{y_i^{m T} y_i^{m+}}{\left\|y_i^m\right\|\left\|y_i^{m +}\right\|\sigma},
\end{equation}
$\theta$ can be selected from traditional CL-based methods, such as CMC, SimCLR, InvP, MoCHi, NNCLR, CO2, etc., representing the definition of positive and negative samples in a manner consistent with them, respectively. $\sigma$ is a scalar temperature parameter.\par 
However, the existing sample-level CL-based methods are used for optimizing the network parameters in the field of deep, they can not obtain the projection matrices to process the traditional multi-view
feature extraction problem. More importantly, even if we construct
InfoNCE loss like (\ref{eq1}) according to these existing methods of defining
positive and negative pairs, and use it to perform feature
extraction, there suffers from the following problems: the above methods promote the consistency of any two cross views from the perspective of the subspace samples, which contains the redundant information suppressing the important discriminative features; these methods do not capture the view-specific discriminative information.\par 

\subsection*{\rm{\textbf{Feature-level CL}}}
To remove the redundant information to achieve the minimality of the subspace discriminative information in accordance with IB principle, the feature-level CL is proposed, which constructs the InfoNCE loss from the perspective of the subspace features. Specifically, the homodimensional subspace features are defined as positive pairs, and the heterodimensional subspace features are defined as negative pairs. The feature-level contrastive loss is as follows:
\begin{equation}
\begin{aligned}\label{eq3}
\mathcal L_{\text{fea}}(P_m,m=1,...,V)=\sum_{m=1}^{V}\sum_{v=1}^{V}\mathbb{E}_{P^k_m, P_v^{k},\left\{P_{v}^{l}\right\}_{l=1}^{d}}\left[-\log \frac{\operatorname{exp}(\operatorname{sim}\left(Y_k^m,  Y_k^v\right)}{\sum_{l=1}^{d} \operatorname{exp}(\operatorname{sim}\left(Y^m_k,  Y_l^{v}\right)}\right],
\end{aligned}
\end{equation}
where 
\begin{equation}
\operatorname{sim}\left(Y_k^m, Y_l^v \right)=\frac{Y_k^m {Y^v_l}^T}{\left\|Y_k^m\right\|\left\|Y^v_l\right\|\sigma},
\end{equation}
$P_m^k, k=1,\cdots,d$ is the $k$th column vector of $P_m$, $Y_k^m=P^{kT}_mX_m$.\par 
We then combine the loss of sample-level CMC and the feature-level contrastive loss as follows:
\begin{equation}\label{e1}
\mathcal L^{\text{CMC}}_{\text {InfoNCE}} +\alpha \mathcal L_{\text{fea}},
\end{equation}
where $\alpha$ is a positive parameter. In the loss (\ref{e1}), the feature-level loss help sample-level loss remove the redundent information in the consistency information.\par

\subsection*{\rm{\textbf{Recovery-level CL}}}
\begin{figure*}[!t]
	\centering
	\subfigure[]{\includegraphics[width=0.24\textwidth]{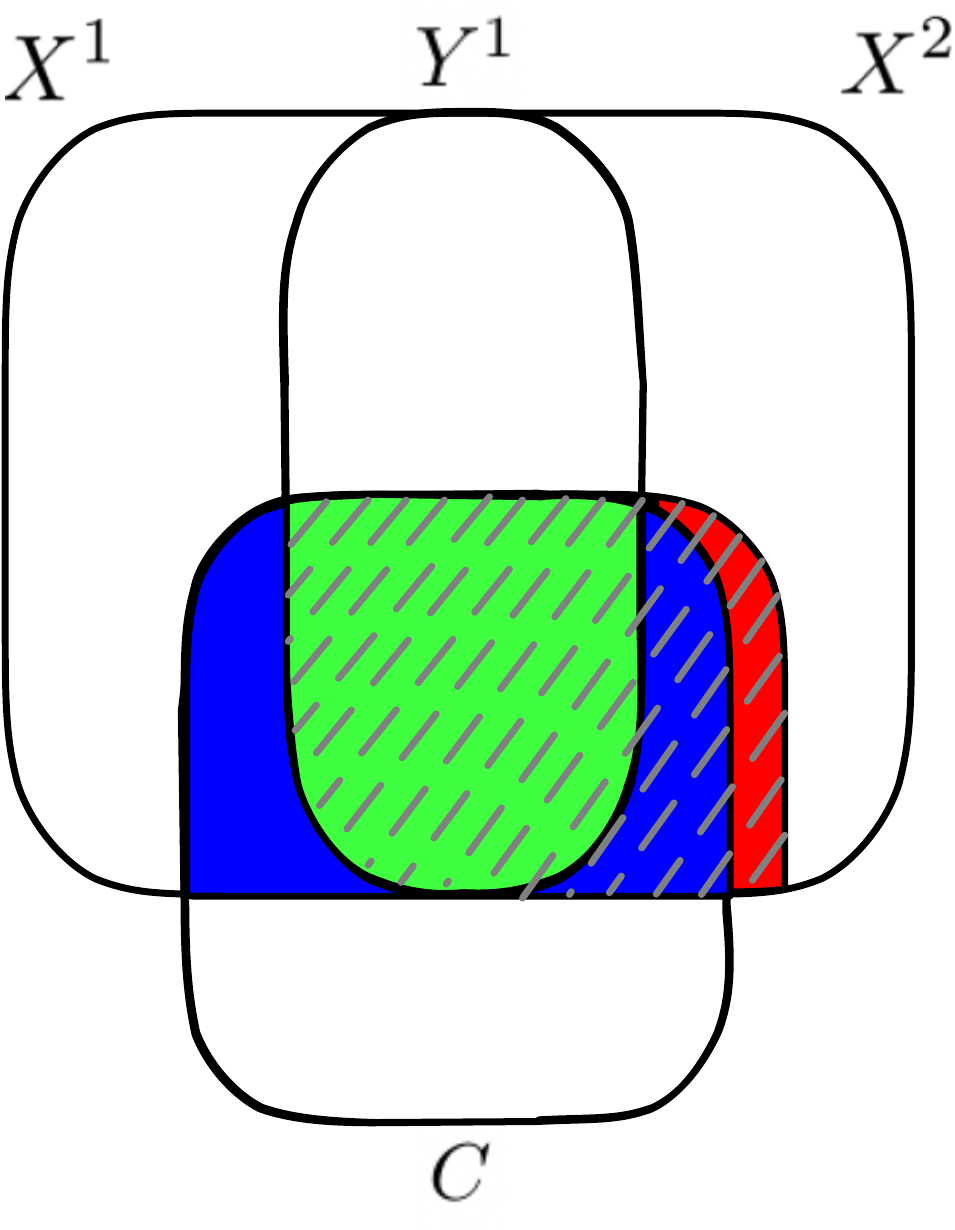}}
	\hspace{1cm}
	\subfigure[]{\includegraphics[width=0.24\textwidth]{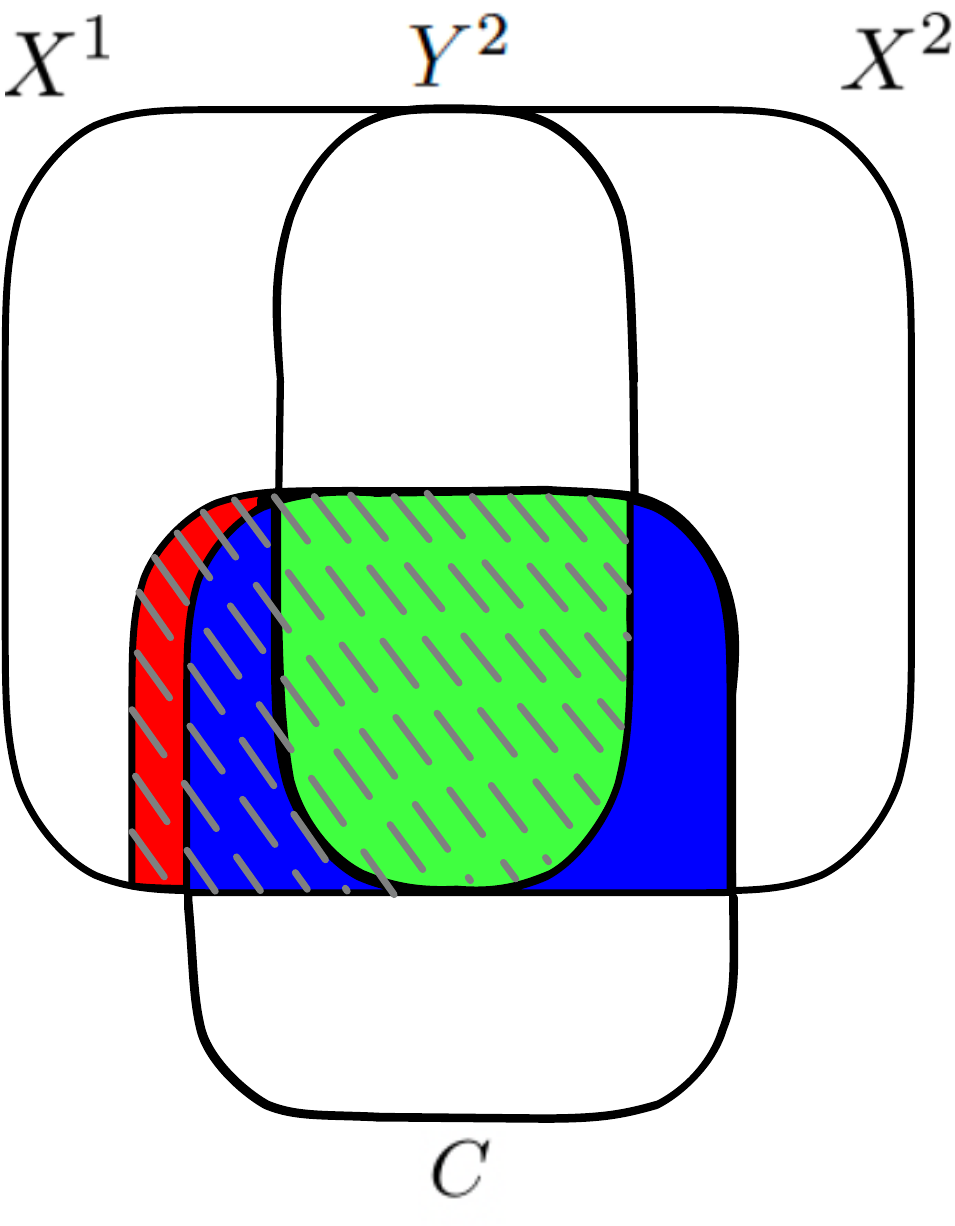}}
	\hspace{1cm}
	\subfigure[]{\includegraphics[width=0.24\textwidth]{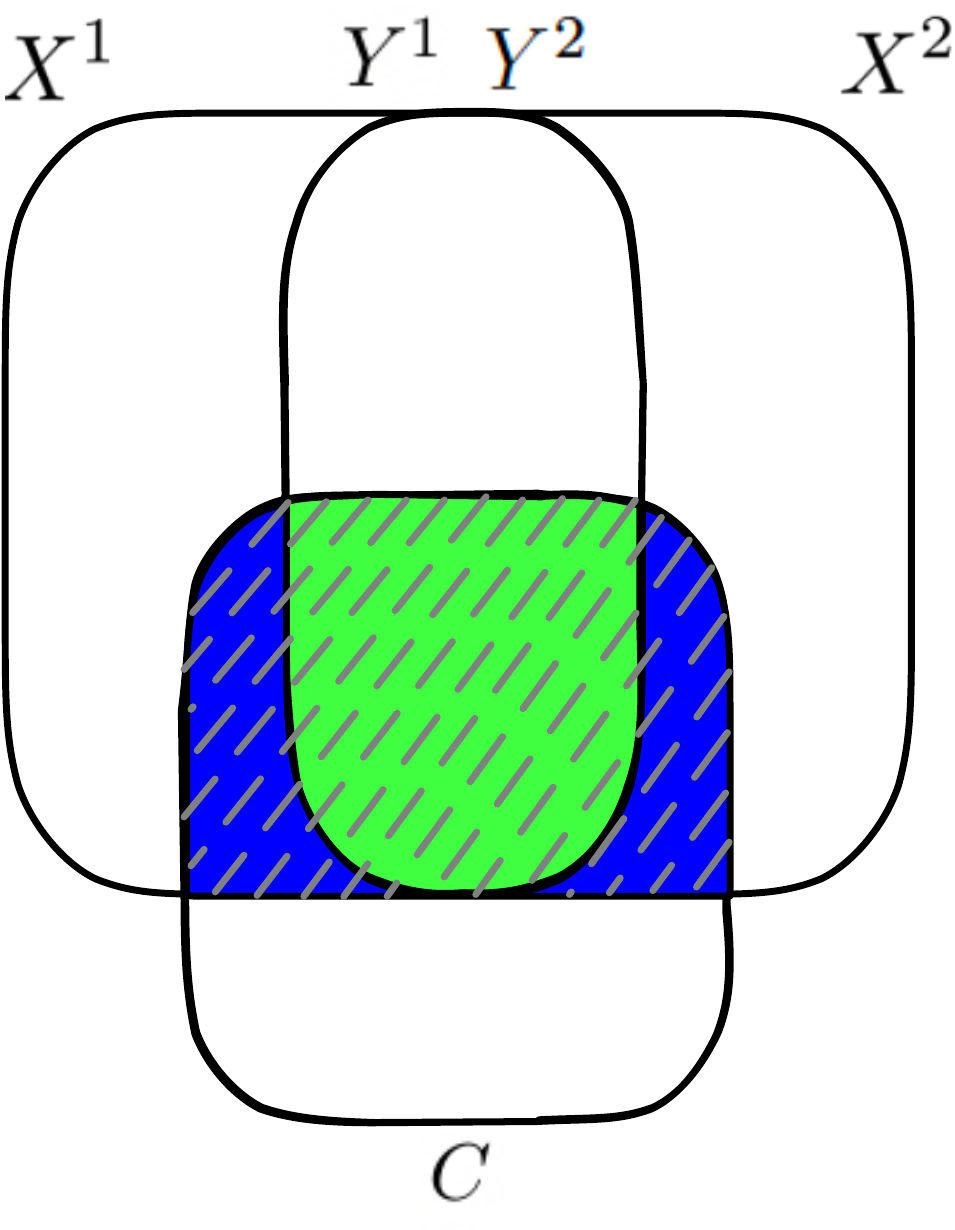}}
	\caption{Illustration of the sufficient yet minimal discriminative information for two views $X^1$ and $X^2$ in terms of information entropy, in which the redundant information is not emphasized. (a) $Y^1$ (shadow area) is obtained through the recovery-level contrastive loss, which contains the view-specific discriminative information in $X^2$.  (b) $Y^2$ (shadow area) is obtained through the recovery-level contrastive loss, which contains the view-specific discriminative information in $X^1$. (c) The sufficient yet minimal discriminative information (shadow area) is obtained by MFETCH.} 
	\label{fig2}
\end{figure*}
To extract the view-specific discriminative information to achieve the sufficiency of the subspace discriminative information in accordance with IB principle, the recovery-level CL is proposed, which constructs the InfoNCE loss from the perspective of the recovered representations. Specifically, the original representations and the recovered subspace representations of same sample are defined as positive pairs, and the original representations and the recovered subspace representations of distinct samples are defined as negative pairs in any two cross views. The recovery-level contrastive loss is as follows:
\begin{equation}
\begin{aligned}\label{eq2}
\mathcal L_{\text{rec}}(P_m,F_m,m=1,...,V)=\sum_{m=1}^{V}\sum_{v=1,v\neq m}^{V}\mathbb{E}_{x_i^m, x_i^{v},\left\{x_{j}^{v}\right\}_{j=1}^{n}}\left[-\log \frac{\operatorname{exp}(\operatorname{sim}\left(x_i^m,  F_m^Ty_i^v\right)}{\sum_{j=1}^{n} \operatorname{exp}(\operatorname{sim}\left(x^m_i,  F_m^Ty_j^{v}\right)}\right],
\end{aligned}
\end{equation}
where 
\begin{equation}
\operatorname{sim}\left(x_i^m, F_m^Ty_j^v \right)=\frac{{x^m_i}^T F_m^Ty^v_j}{\left\|x^m_i\right\|\left\|F_m^Ty^v_j\right\|\sigma},
\end{equation}
the matrices $F_m\in R^{d\times D_m}, m=1,2,...,V$ derive the recovered representations of the $m$th view. In fact, if a subspace view can be recovered to another original views, then it means that the subspace view contains as much the view-specific discriminative information of original views as possible. A more intuitively illustration is shown in Figure \ref{fig2}(a) and (b). \\

\subsection*{\rm{\textbf{MFETCH}}}
Although the recovery-level contrastive loss can capture the view-specific discriminative information, which inevitably contains the view-specific non-discriminative information, as shown in the red areas of Figure \ref{fig2}(a) and (b). In MFETCH, we combine the sample-, feature- and recovery-level contrastive losses to promote the consistency of any two cross views while capturing the view-specific discriminative information, so as to remove the view-specific non-discriminative information and the redundent information.  A more intuitively illustration is shown in Figure \ref{fig2}(c). The specific loss of MFETCH is as follows:
\begin{equation}\label{eq4}
\mathcal L=\mathcal L^{\text{CMC}}_{\text {InfoNCE}} +\alpha \mathcal L_{\text{fea}}+\beta \mathcal L_{\text{rec}},
\end{equation}
where $\beta$ is a positive parameters.\par
In fact, the view-specific discriminative information also contains the redundant information, but feature-level contrastive loss can solve this problem naturally, so the redundant information is not shown in Figure \ref{fig2}.
Moreover, the balance parameters $\alpha$ and $\beta$ are particularly important for extracting the sufficient yet minimal subspace discriminative information in accordance with IB principle more accurately.  
In addition, the feature- and recovery-level contrastive losses can also be combined with other sample-level contrastive loss, such as SimCLR, InvP, MoCHi, NNCLR, CO2, PCL, HCSC, etc., which provides a new framework for future CL.\par \par 

\subsection{Theoretical Analysis}
The relationship between MFEDCH and mutual information are analyzed from the theoretical aspect. For convenience, we make $z_j^v=F_m^Ty_j^v, v=1,\cdots,V,j=1,\cdots,n$.
Obviously, when $m$ and $v (m\neq v)$ are fixed, $z^v_j$ is a positive sample of $x^m_i$ iff $i=j$, otherwise $z^v_j$ is a negative sample of $x^m_i$. Naturally, the probability that the sample $z^v_j$ is a positive sample of $x^m_i$ is equal to $p(i=j |z^v_j, x^m_i)$, and the recovery-level loss is equivalent to:
\begin{equation}
\begin{aligned}
\mathcal L_{\text{rec}}=&\sum_{m=1}^{V}\sum_{v=1,v\neq m}^{V}\mathbb{E}_{x_i^m, z_j^{v}}\left[-\log   p(i=j|z^v_j, x^m_i)\right]\\
\end{aligned}
\end{equation}
Therefore, there are two prior distributions $p(i=j) = \frac{1}{n}$ and $p(i\neq j) = \frac{n-1}{n}$. According to the Bayesian formula, the following derivation is made:
\begin{equation}
\begin{aligned}
p(i=j|z^m_j, x^v_i)&=\frac{p(z^v_j, x^m_i|i=j)p(i=j)}{p(z^v_j, x^m_i|i=j)p(i=j)+p(z^v_j, x^m_i|i\neq j)p(i\neq j)}\\
&=\frac{p(z^v_j, x^m_i|i=j)\frac{1}{n}}{p(z^v_j, x^m_i|i=j)\frac{1}{n}+p(z^v_j, x^m_i|i\neq j)\frac{n-1}{n}}\\
&=\frac{p(z^v_j, x^m_i|i=j)}{p(z^v_j, x^m_i|i=j)+(n-1)p(z^v_j, x^m_i|i\neq j)}\\
&=\frac{p(z^v_j, x^m_i)}{p(x^v_j, x^m_i)+(n-1)p(z^v_j)p(x^m_i)}.\\
\end{aligned}
\end{equation}
Further derivation, there is the following formula:
\begin{equation}
\begin{aligned}
\mathcal L_{\text {rec}}&\geq\sum_{m=1}^{V}\sum_{v=1,v\neq m}^{V}\mathbb{E}_{x_i^m, z_j^{v}}\left[-\log   p(i=j|z^v_j, x^m_i)\right]\\
&=\sum_{m=1}^{V}\sum_{v=1,v\neq m}^{V}\mathbb{E}_{x_i^m, z_j^{v}}\left[-\log  \frac{p(z^v_j, x^m_i)}{p(z^v_j, x^m_i)+(n-1)p(z^v_j)p(x^m_i)}\right]\\
&=\sum_{m=1}^{V}\sum_{v=1,v\neq m}^{V}\mathbb{E}_{x_i^m, z_j^{v}}\left[\log  \frac{p(z^v_j, x^m_i)+(n-1)p(z^v_j)p(x^m_i)}{p(z^v_j, x^m_i)}\right]\\
&=\sum_{m=1}^{V}\sum_{v=1,v\neq m}^{V}\mathbb{E}_{x_i^m, z_j^{v}}\log  \left[1+\left( n-1\right) \frac{p(z^v_j)p(x^m_i)}{p(z^v_j, x^m_i)}\right]\\
&=\sum_{m=1}^{V}\sum_{v=1,v\neq m}^{V}\mathbb{E}_{x_i^m, z_j^{v}}\log \left[ \frac{p(z^v_j,x^m_i)-p(z^v_j)p(x^m_i)}{p(z^v_j,x^m_i)}+ n \frac{p(z^v_j)p(x^m_i)}{p(z^v_j,x^m_i)}\right]\\
\end{aligned}
\end{equation}
Since $x^m_i$ and $z^v_j$ are positive samples, $x^m_i$ and $z^v_j$ are not independent, then $p(z^v_j,x^m_i)-p(z^v_j)p(x^m_i)>0$. Therefore, we can get the following derivation
\begin{equation}
\begin{aligned}
\mathcal L_{\text {str}}&\geq\sum_{m=1}^{V}\sum_{v=1,v\neq m}^{V}\mathbb{E}_{x_i^m, z_j^{v}}\log \left[ \frac{p(z^v_j,x^m_i)-p(z^v_j)p(x^m_i)}{p(z^v_j,x^m_i)}+ n \frac{p(z^v_j)p(x^m_i)}{p(z^v_j,x^m_i)}\right]\\
&\geq\sum_{m=1}^{V}\sum_{v=1,v\neq m}^{V}\mathbb{E}_{x_i^m, z_j^{v}}\log \left[ n \frac{p(z^v_j)p(x^m_i)}{p(z^v_j,x^m_i)}\right]\\
&= \sum_{m=1}^{V}\sum_{v=1,v\neq m}^{V}\left[\log (n)-I(z^v_j,x^m_i)\right] 
\end{aligned}
\end{equation}
where $I(z^v_j,x^m_i)$ represents the mutual information between $z^v_j$ and $x^m_i$.
Therefore, we can get $-\mathcal L_{\text {rec}}\leq \sum_{m=1}^{V}\sum_{v=1,v\neq m}^{V}\left[I(z^v_j,x^m_i)-log\left(n\right)\right]$, and minimizing $\mathcal L_{\text {rec}}$ is equivalent to maximizing the mutual information of all positive pairs. In fact, the recovery-level contrastive loss actually maximizes the mutual information of the original and recovered representations of same sample in any cross views, which captures the nonlinear statistical dependencies between the original and recovered representations of same sample in any two cross views, and thus, can serve as a measure of true dependence. Similarly, we can also demonstrate the feature-level contrastive actually maximizes the mutual information of the subspace homodimensional features and minimizes the mutual information of the subspace heterodimensional features, and thus, can remove the redundant information more effectively.\par
\subsection{Optimization Solution} 
\uppercase\expandafter{\romannumeral 1}. When $P_m,m=1,...,V$ are fixed, the loss function (\ref{eq4}) becomes:
\begin{equation}\label{F}
\mathcal L(F_m)=\sum_{m=1}^{V}\sum_{v=1,v\neq m}^{V}\mathbb{E}_{x_i^m, x_i^{v},\left\{x_{j}^{v}\right\}_{i=1}^{n}}\left[-\log \frac{\operatorname{exp}(\operatorname{sim}\left(x_i^m,  F_m^TP_v^T\tilde{x}_i^v\right)}{\sum_{j=1}^{n} \operatorname{exp}(\operatorname{sim}\left(x^m_i,  F_m^TP_v^T\tilde{x}_j^{v}\right)}\right].
\end{equation}
The derivative of (\ref{F}) can be calculated as follows:
\begin{equation}\label{equ5}
\begin{aligned}
\nabla \mathcal L(F_m)=
&\sum_{v=1,v\neq m}^{V} \mathbb{E}_{x_i^m, x_i^{v},\left\{x_{j}^{v}\right\}_{i=1}^{n}}\left[-\frac{\sum_{j=1}^{n} \operatorname{exp}(\operatorname{sim}\left(x^m_i,  F_m^TP_v^Tx_j^{v}\right)}{\operatorname{exp}(\operatorname{sim}\left(x_i^m,  F_m^TP_v^Tx_i^v\right)}\right] \cdot\\
&\left\{\operatorname{exp}(\operatorname{sim}\left(x_i^m,  F_m^TP_v^Tx_i^v\right) \cdot \triangle^{m v}_{i i} \cdot \sum_{j=1}^{n} \operatorname{exp}(\operatorname{sim}\left(x^m_i,  F_m^TP_v^Tx_j^{v}\right)-\right. \\
&\left.\sum_{j=1}^{n}\left[\operatorname{exp}(\operatorname{sim}\left(x^m_i,  F_m^TP_v^Tx_j^{v}\right) \cdot \triangle^{m v}_{i j}\right] \cdot \operatorname{exp}(\operatorname{sim}\left(x_i^m,  F_m^TP_v^Tx_i^v\right) \right\}\bigg/\\
&{\left[\sum_{j=1}^{n}\operatorname{exp}(\operatorname{sim}\left(x^m_i,  F_m^TP_v^Tx_j^{v}\right)\right]^{2}},\\
\end{aligned}
\end{equation}
where
\begin{equation}
\begin{aligned}
\triangle_{i j}^{m v}= 
&\left\{P_v^Tx_{j}^{v}x^{m T}_i \cdot\left\|x_{i}^{m}\right\|\left\|F_m^TP_v^Tx_{j}^{v}\right\| \sigma_2\right. \\
&\left.-\left(x_{j}^{v T} P_vF_mF_m^T P_v^{T} x_{j}^{v}\right)^{-\frac{1}{2}} \cdot P_v^Tx_{j}^{v}x_{j}^{v T}P_vF_m\cdot \left\|x_{i}^{m}\right\|\sigma_2\right\} \bigg/ \\
&\left(\left\|x_{i}^{m}\right\|\left\|F_m^TP_v^Tx_{j}^{v}\right\| \sigma_2\right)^{2}.
\end{aligned}
\end{equation}
\uppercase\expandafter{\romannumeral 2}. The loss function (\ref{eq1}) can be transformed into:
\begin{equation}\label{P1}
\mathcal L^{\text{CMC}}_{\text{InfoNCE}}(P)=\sum_{m=1}^{V}\sum_{v=1,v\neq m}^{V}\mathbb{E}_{\tilde{x}_i^m, \tilde{x}_i^{v},\left\{\tilde{x}_{j}^{v}\right\}_{i=1}^{n}}\left[-\log \frac{\operatorname{exp}(\operatorname{sim}\left(P^T\tilde{x}_i^m,  P^T\tilde{x}_i^v\right)}{\sum_{j=1}^{n} \operatorname{exp}(\operatorname{sim}\left(P^T\tilde{x}^m_i,  P^T\tilde{x}_j^{v}\right)}\right],
\end{equation}
where
\begin{equation}
P=[P_1;...;P_V]\in R^{D\times d},
\end{equation}
\begin{equation}
\tilde{X}^m=[\underbrace{\mathbf{0};...;\mathbf{0}}_{m-1};X^{m};\underbrace{\mathbf{0};...;\mathbf{0}}_{V-m}]\in R^{D\times n},
\end{equation}
$v$-th $\mathbf{0}$ is the zero matrix of the corresponding scale of the data in the $v$-th view, $\tilde{x}^m_i$ is the $i$-th column vector of $\tilde{X}^m$, $D=\sum_{m=1}^{n}D_m$. The derivative of (\ref{P1}) can be calculated as follows:
\begin{equation}\label{grad1}
\begin{aligned}
\nabla \mathcal L^{\text{CMC}}_{\text{InfoNCE}}(P)=
&\sum_{m=1}^{V} \sum_{v=1}^{V} \mathbb{E}_{\tilde{x}_{i}^{m}, \tilde{x}_{i}^{v},\left\{\tilde{x}_{j}^{v}\right\}_{i=1}^{n}}\left[-\frac{\sum_{j=1}^{n} \exp \left(\operatorname{sim}\left(P^{T} \tilde{x}_{i}^{m}, P^{T} \tilde{x}_{j}^{v}\right)\right)}{\exp \left(\operatorname{sim}\left(P^{T} \tilde{x}_{i}^{m}, P^{T} \tilde{x}_{i}^{v}\right)\right)}\right] \cdot\\
&\left\{\exp \left(\operatorname{sim}\left(P^{T} \tilde{x}_{i}^{m}, P^{T} \tilde{x}_{i}^{v}\right)\right) \cdot \nabla 1^{m v}_{i i} \cdot \sum_{j=1}^{n} \exp \left(\operatorname{sim}\left(P^{T} \tilde{x}_{i}^{m}, P^{T} \tilde{x}_{j}^{v}\right)\right)-\right. \\
&\left.\sum_{j=1}^{n}\left[\exp \left(\operatorname{sim}\left(P^{T} \tilde{x}_{i}^{m}, P^{T} \tilde{x}_{j}^{v}\right)\right) \cdot \nabla 1^{m v}_{i j}\right] \cdot \exp \left(\operatorname{sim}\left(P^{T} \tilde{x}_{i}^{m}, P^{T} \tilde{x}_{i}^{v}\right)\right) \right\}\bigg/\\
&{\left[\sum_{j=1}^{n} \exp \left(\operatorname{sim}\left(P^{T} \tilde{x}_{i}^{m}, P^{T} \tilde{x}_{j}^{v}\right)\right)\right]^{2}},\\
\end{aligned}
\end{equation}
where
\begin{equation}
\begin{aligned}
\nabla 1_{i j}^{m v}= 
&\left\{\left(\tilde{x}_{i}^{m} \tilde{x}_{j}^{v T}+\tilde{x}_{j}^{v} \tilde{x}_{i}^{m T}\right) P \cdot\left\|P^{T} \tilde{x}_{i}^{m}\right\|\left\|P^{T} \tilde{x}_{j}^{v}\right\| \sigma_1\right. \\
&-\left[\left(\tilde{x}_{i}^{m T} P P^{T} \tilde{x}_{i}^{m}\right)^{-\frac{1}{2}} \cdot \tilde{x}_{i}^{m} \tilde{x}_{i}^{m T} P \cdot\left\|P^{T} \tilde{x}_{j}^{v}\right\| \sigma_1\right. \\
&\left.\left.+\left(\tilde{x}_{j}^{v T} P P^{T} \tilde{x}_{j}^{v}\right)^{-\frac{1}{2}} \cdot \tilde{x}_{j}^{v} \tilde{x}_{j}^{v T} P \cdot\left\|P^{T} \tilde{x}_{i}^{m}\right\| \sigma\right] \cdot \tilde{x}_{i}^{m T} P P^{T} \tilde{x}_{j}^{v}\right\} \bigg/ \\
&\left(\left\|P^{T} \tilde{x}_{i}^{m}\right\|\left\|P^{T} \tilde{x}_{j}^{v}\right\| \sigma_1\right)^{2}.
\end{aligned}
\end{equation}
When $F^m,m=1,...,V$ are fixed, the loss function (\ref{eq2}) becomes:
\begin{equation}\label{P2}
\mathcal L_{\text{rec}}(P)=\sum_{m=1}^{V}\sum_{v=1,v\neq m}^{V}\mathbb{E}_{x_i^m, x_i^{v},\left\{x_{j}^{v}\right\}_{i=1}^{n}}\left[-\log \frac{\operatorname{exp}(\operatorname{sim}\left(x_i^m,  F_m^TP^T\tilde{x}_i^v\right)}{\sum_{j=1}^{n} \operatorname{exp}(\operatorname{sim}\left(x^m_i,  F_m^TP^T\tilde{x}_j^{v}\right)}\right].
\end{equation}
The derivative (\ref{P2}) can be calculated as follows:
\begin{equation}
\begin{aligned}
\nabla \mathcal L_{\text{rec}}(P)=
&\sum_{m=1}^{V} \sum_{v=1,v\neq m}^{V} \mathbb{E}_{x_i^m, x_i^{v},\left\{x_{j}^{v}\right\}_{i=1}^{n}}\left[-\frac{\sum_{j=1}^{n} \operatorname{exp}(\operatorname{sim}\left(x^m_i,  F_m^TP^T\tilde{x}_j^{v}\right)}{\operatorname{exp}(\operatorname{sim}\left(x_i^m,  F_m^TP^T\tilde{x}_i^v\right)}\right] \cdot\\
&\left\{\operatorname{exp}(\operatorname{sim}\left(x_i^m,  F_m^TP^T\tilde{x}_i^v\right) \cdot \nabla 2^{m v}_{i i} \cdot \sum_{j=1}^{n} \operatorname{exp}(\operatorname{sim}\left(x^m_i,  F_m^TP^T\tilde{x}_j^{v}\right)-\right. \\
&\left.\sum_{j=1}^{n}\left[\operatorname{exp}(\operatorname{sim}\left(x^m_i,  F_m^TP^T\tilde{x}_j^{v}\right) \cdot \nabla 2^{m v}_{i j}\right] \cdot \operatorname{exp}(\operatorname{sim}\left(x_i^m,  F_m^TP^T\tilde{x}_i^v\right) \right\}\bigg/\\
&{\left[\sum_{j=1}^{n}\operatorname{exp}(\operatorname{sim}\left(x^m_i,  F_m^TP^T\tilde{x}_j^{v}\right)\right]^{2}},\\
\end{aligned}
\end{equation}
where
\begin{equation}
\begin{aligned}
\nabla 2_{i j}^{m v}= 
&\left\{\tilde{x}_{j}^{v}x^{m T}_iF_m^T \cdot\left\|x_{i}^{m}\right\|\left\|F_m^TP^T\tilde{x}_{j}^{v}\right\| \sigma_2\right. \\
&\left.-\left(\tilde{x}_{j}^{v T} PF_mF_m^T P^{T} \tilde{x}_{j}^{v}\right)^{-\frac{1}{2}} \cdot \tilde{x}_{j}^{v}\tilde{x}_{j}^{v T}PF_mF_m^T\cdot \left\|x_{i}^{m}\right\|\sigma_2\right\} \bigg/ \\
&\left(\left\|x_{i}^{m}\right\|\left\|F_m^TP^T\tilde{x}_{j}^{v}\right\| \sigma_2\right)^{2}.
\end{aligned}
\end{equation}
Furthermore, the derivative of (\ref{eq3}) can be calculated as follows:
\begin{equation}
\nabla \mathcal L_{\text{fea}}(P)=[\nabla P^1_1,...,\nabla P^d_1;...;\nabla P^1_V,...,\nabla P^d_V],
\end{equation}
where
\begin{equation}
\begin{aligned}
\nabla P_m^k=
&\sum_{m=1}^{V} \sum_{v=1}^{V} \mathbb{E}_{P_m^k, P_v^k,\left\{P_{l}^{v}\right\}_{l=1}^{d}}\left[-\frac{\sum_{l=1}^{d} \exp \left(\operatorname{sim}\left(X^{mT}P^k_m, X^{v T}P^l_v\right)\right)}{\exp \left(\operatorname{sim}\left(X^{mT}P^k_m, X^{v T}P^k_v\right)\right)}\right] \cdot\\
&\left\{\exp \left(\operatorname{sim}\left(X^{mT}P^k_m, X^{v T}P^k_v\right)\right) \cdot \nabla 3^{m v}_{i i} \cdot \sum_{l=1}^{d} \exp \left(\operatorname{sim}\left(X^{mT}P^k_m, X^{v T}P^l_v\right)\right)-\right. \\
&\left.\sum_{j=1}^{n}\left[\exp \left(\operatorname{sim}\left(X^{mT}P^k_m, X^{v T}P^l_v\right)\right) \cdot \nabla 3^{m v}_{i j}\right] \cdot \exp \left(\operatorname{sim}\left(X^{mT}P^k_m, X^{v T}P^k_v\right)\right) \right\}\bigg/\\
&{\left[\sum_{j=1}^{n} \exp \left(\operatorname{sim}\left(P^{T} \tilde{x}_{i}^{m}, P^{T} \tilde{x}_{j}^{v}\right)\right)\right]^{2}},\\
\end{aligned}
\end{equation}

\begin{equation}
\begin{aligned}
\nabla 3_{i j}^{m v}= 
&\left\{X^{m} X^{v T} P_v^k\cdot\left\|X^{mT}P_m^k\right\|\left\|X^{vT}P_v^l\right\| \sigma_3\right. \\
&\left.\left(P_m^{kT}X^mX^{m T}P_m^k\right)^{-\frac{1}{2}}\cdot X^mX^{m T}P_m^k\cdot \left\|X^{v T}P_v^l\right\|\cdot P^k_mX^mX^vP_v^l \right\} \bigg/ \\
&\left(\left\|X^{mT}P_m^k\right\|\left\|X^{vT}P_v^l\right\| \sigma_3\right)^{2}.
\end{aligned}
\end{equation}
Therefore, the derivative of (\ref{eq4}) can be calculated as follows:
\begin{equation}\label{equ4}
\nabla \mathcal L(P)=\nabla \mathcal L_{\text{InfoNCE}}^{\text{CMC}}(P)+\nabla \mathcal L_{\text{rec}}(P)+\nabla \mathcal L_{\text{fea}}(P).
\end{equation}
The problem (\ref{eq4}) is solved by using the Adam optimizer. In Adam, the parameters $\gamma$, $\beta_1$, $\beta_2$, and $\epsilon$ represent the learning rate, the exponential decay rate of the first- and second-order moment estimation, and the parameter to prevent division by zero in the implementation, respectively. The optimization algorithm for problem (\ref{eq4}) is summarized in the Algorithm 1. The convergent condition used in our experiments
is set as $\left|\mathcal L(P_m^t, m=1,..,V)-\mathcal L(P_m^t+1, m=1,...,V)\right|\leq 10^{-3}$.
\begin{algorithm}[!h]\label{Algorithm1}
	\caption{MFETCH} 
	{\bf Input:} 
	
	Data matrix: $X^m\in R^{D_m\times n}, m=1,...,V$, $d$, $\sigma, \alpha, \beta$, $\gamma$, $\beta_1$, $\beta_2$, $\epsilon$.\\
	$t= 0$ (Initialize number of iterations)\\
	$P_m^0$ (Initialize projection matrices)\\
	\hspace*{0.02in} {\bf Output:} 
	Projection matrix $P_m$
	\begin{algorithmic}
		\WHILE{$P^t_m, m=1,...,V$ not converged} 
		\FOR{$m=1$ to $V$}
		\STATE
		Updata $F_m^t$ using Adam algorithm by (\ref{equ5})\\
		\ENDFOR\\
		Updata $P^t_m, m=1,...,V$ using Adam algorithm by (\ref{equ4})\\
		$t=t+1$\\
		\ENDWHILE
		\RETURN $P^t_m, m=1,...,V$
	\end{algorithmic}
\end{algorithm}

\section{Experiments}
\subsection{Data Descriptions}
In the numerical experiments, we used four real-world datasets to
demonstrate the performance advantages of the proposed MFEDCH.\par 
Yale dataset: The dataset is created by Yale University Computer Vision and Control Center, containing data of $15$ individuals, wherein each person has 11 frontal images captured under various lighting conditions. \par 
Multiple features (MF) dataset: The dataset consists of six different types of features extracted by the handwritten numbers from 0 to 9. It contains 2000 samples of 10 objects with 200 instances per subject. \par 
In our experiments, GS (gray-scale intensity, 256 features) and LBP (local binary pattern histogram, 256 features) are applied to
two views for Yale dataset. FAC (profile correlations, 216 features), FOU (Fourier coefficients, 76 features), and PIX (pixel averages in $2\times3$ windows, 240 features) are applied to three views for MF dataset. 
\subsection{Comparison methods}
To demonstrate the effectiveness of the proposed MFETCH, we compared the performance of it with the traditional multi-view feature extraction methods LPCCA, ALPCCA\cite{alpcca}, GDMCCA\cite{gmcca}, and KMSA-PCA\cite{kmsa}, as well as the traditional deep learning methods CMC and Barlow Twins.\par
LPCCA is a typical two-view feature extraction method which aims at preserving local neighbor information of the samples in each views.\par 
ALPCCA is a typical two-view feature extraction which aims at preserving local neighbor information of the samples in cross views.\par 
GDMCCA is a typical multi-view feature extraction method which aims at preserving common graph structure of the samples.\par
KMSA-PCA is a new multi-view feature extraction method which aims at exploring graph structure information in kernel space.\par
CMC is a typical deep learning method based on CL that defines the same samples as positive pairs and distinct samples as negative pairs in any two cross views. Its loss function is used
to obtain the projection matrices of performing multi-view  feature extraction.\par 
Barlow Twins is a new deep learning method that promotes the consistency of subspace features in any two cross views.  Its loss function is used
to obtain the projection matrices of performing multi-view  feature extraction.\par 
\subsection{Experiments setups}
\begin{table}[!ht]      
	\centering
	\caption{Experimental details about datasets.}
	\label{T01}
	\resizebox{1\textwidth}{!}{
		\begin{tabular}[t]{c c c c c}
			\hline
			Dataset & View &No. of classes &No. of samples &No. of training samples\\
			\hline
			Yale&GS, LBP&15&165&$M$=4,6,8\\
			MF&FAC, FOU, PIX&10&2000&$M$=6,8,10\\
			\hline
		\end{tabular}} 
	\end{table}	
	We demonstrate the strong performance on classification tasks to assess the effectiveness of our proposed MFETCH. The $k$-nearest neighbor classifier ($k$=1) was used in the experiment. Moreover, $M$ samples of each class from Yale, ORL, COIL20, and MF datasets were randomly selected for training, whereas the remaining data were used for testing. The details are presented in Table \ref{T01}.  All processes are repeated five times, and the final evaluation criteria constitute the classification accuracy oflow-dimensional features obtained by \uppercase\expandafter{\romannumeral1} and \uppercase\expandafter{\romannumeral2}. The experiments are implemented using MATLAB R2018a on a computer with an Intel Core i5-9400 2.90 GHz CPU and Windows 10 operating system.\par 
	\uppercase\expandafter{\romannumeral1}. low-dimensional features for each view: 
	\begin{equation}\label{1}
	Y_{m}=P_{m}^{\mathrm{T}} X_{m}, m=1, \ldots, V
	\end{equation}
	
	\uppercase\expandafter{\romannumeral2}. low-dimensional features by fusion strategy:
	\begin{equation}\label{2}
	Y=P_{1}^{\mathrm{T}} X_{1}+\cdots+P_{V}^{\mathrm{T}} X_{V}
	\end{equation}
	
	In addition, the performance of various feature extraction methods was evaluated by setting certain parameters in advance. First, the appropriate default parameters for testing machine-learning problems in the Adam optimizer are $\gamma = 0.001, \beta_1 = 0.9, \beta_2 = 0.999$, and $\epsilon = 10^{-8}$. In MFETCH, the balance parameters $\alpha$ and $\beta$ are set to $1$, whereas the temperature parameters $\sigma$ is set to $0.1$. 
	\subsection{Analysis of results}
	\begin{table*}
		\caption{Experimental results of Yale dataset (maximum classification accuracy $\pm$ standard deviations \%).}
		\label{T1}
		\resizebox{\textwidth}{!}{
			\begin{tabular}{c c c c c c c c}
				\toprule [2pt]
				View&LPCCA& ALPCCA & GDMCCA & KMSA-PCA& CMC & Barlow Twins & MFETCH\\
				\hline
				\hline
				\multicolumn{8}{c}{\textbf{Train-4}}\\
				\hline
				\textbf{GS}  
				&$66.19\pm1.99$&$63.24\pm3.13$&$73.52\pm2.37$&$73.33\pm0.67$&$78.24\pm0.80$&$77.14\pm2.43$&\bm{$79.62\pm1.09$}\\
				\textbf{LBP}
				&$55.67\pm9.05$&$65.52\pm5.87$&$69.52\pm6.70$&$80.57\pm7.30$&$84.33\pm4.60$&$83.81\pm4.62$&\bm{$84.76\pm4.26$}\\
				\textbf{Mean}	
				&$61.43\pm5.52$&$64.38\pm4.50$&$71.52\pm4.54$&$76.95\pm3.99$&$81.19\pm3.03$&$80.48\pm3.53$&\bm{$82.19\pm2.68$}\\
				\textbf{\uppercase\expandafter{\romannumeral2}} 
				&$64.38\pm6.79$&$69.48\pm5.87$&$74.48\pm5.28$&$75.05\pm4.28$&$77.10\pm6.21$&$77.90\pm4.23$&\bm{$78.29\pm4.64$}\\
				\hline
				\hline
				
				\multicolumn{8}{c}{\textbf{Train-6}}\\
				\hline
				\textbf{GS}  
				&$63.33\pm2.40$&$64.67\pm9.09$&\bm{$79.20\pm3.96$}&$73.07\pm4.36$&$75.62\pm1.23$&$76.27\pm4.46$&$78.60\pm1.98$\\
				\textbf{LBP}
				&$64.00\pm6.53$&$67.07\pm7.02$&$68.00\pm1.89$&$83.20\pm3.84$&$89.11\pm1.31$&$90.40\pm3.45$&\bm{$91.93\pm1.12$}\\		
				\textbf{Mean}	
				&$63.67\pm4.47$&$65.87\pm8.06$&$73.60\pm2.93$&$78.13\pm4.10$&$82.36\pm1.27$&$83.33\pm3.96$&\bm{$85.13\pm1.38$}\\
				\textbf{\uppercase\expandafter{\romannumeral2}} 
				&$67.60\pm4.91$&$76.27\pm3.67$&$79.47\pm4.28$&$74.93\pm4.36$&$80.95\pm1.33$&$77.20\pm2.56$&\bm{$81.13\pm0.94$}\\
				\hline
				\hline
				
				\multicolumn{8}{c}{\textbf{Train-8}}\\
				\hline
				\textbf{GS}  
				&$66.22\pm7.27$&$76.00\pm3.65$&\bm{$80.44\pm5.75$}&$80.00\pm7.86$&$77.22\pm4.82$&$76.44\pm5.12$&$78.22\pm4.82$\\
				\textbf{LBP}
				&$79.56\pm2.90$&$76.44\pm9.07$&$75.56\pm4.97$&$86.67\pm7.03$&$89.67\pm4.82$&$89.78\pm4.04$&\bm{$93.33\pm5.44$}\\		
				\textbf{Mean}	
				&$72.89\pm5.09$&$76.22\pm6.36$&$78.00\pm5.36$&$83.33\pm7.45$&$83.44\pm4.82$&$83.11\pm4.58$&\bm{$85.78\pm5.13$}\\
				\textbf{\uppercase\expandafter{\romannumeral2}}
				&$76.89\pm4.33$&$82.67\pm3.65$&\bm{$85.33\pm4.33$}&$80.44\pm6.36$&$80.00\pm5.67$&$77.33\pm3.98$&$81.33\pm5.35$\\
				\bottomrule[2pt]
			\end{tabular}}
		\end{table*}

				\begin{table*}
					\caption{Experimental results of MF dataset (maximum classification accuracy $\pm$ standard deviations \%).}
					\label{T4}
					\resizebox{\textwidth}{!}{
						\begin{tabular}{c c c c c c c c}
							\toprule [2pt]
							View& LPCCA & ALPCCA & GDMCCA & KMSA-PCA & CMC & Barlow Twins & MFEDCH\\
							\hline
							\hline
							\multicolumn{8}{c}{\textbf{Train-6}}\\
							\hline
							\textbf{FAC}  
							&$54.37\pm3.77$&$65.78\pm2.00$&$74.62\pm1.25$&$84.38\pm2.76$&$85.85\pm0.71$&$86.06\pm1.77$&\bm{$87.13\pm0.67$}\\
							\textbf{FOU}
							&$56.02\pm4.14$&$54.93\pm3.79$&$68.38\pm5.32$&$76.19\pm2.05$&$77.74\pm3.31$&$76.82\pm3.13$&\bm{$78.61\pm3.27$}\\
							\textbf{PIX}	
							&$73.47\pm2.25$&$74.16\pm1.30$&$74.30\pm1.86$&$73.13\pm3.85$&$73.37\pm1.54$&$74.62\pm1.82$&\bm{$74.69\pm1.85$}\\	
							\textbf{Mean}	
							&$61.29\pm3.39$&$64.96\pm2.36$&$72.43\pm2.81$&$77.90\pm2.89$&$78.98\pm1.73$&$79.17\pm2.24$&\bm{$80.12\pm2.03$}\\	
							\textbf{\uppercase\expandafter{\romannumeral2}}
							&$63.26\pm3.82$&$66.94\pm4.01$&\bm{$79.91\pm1.10$}&$76.18\pm2.05$&$77.66\pm3.36$&$77.02\pm3.19$&$78.57\pm3.24$\\
							\hline
							\hline
							
							\multicolumn{8}{c}{\textbf{Train-8}}\\
							\hline
							\textbf{FAC}   
							&$66.87\pm2.94$&$70.43\pm1.59$&$78.10\pm2.09$&$86.75\pm1.88$&$88.56\pm1.69$&$88.59\pm1.16$&\bm{$89.63\pm1.82$}\\
							\textbf{FOU}
							&$54.39\pm2.80$&$51.73\pm1.87$&$66.77\pm1.86$&$78.17\pm3.63$&$79.82\pm3.43$&$79.29\pm3.42$&\bm{$81.07\pm3.75$}\\
							\textbf{PIX}	
							&$71.80\pm2.33$&$75.18\pm3.01$&$74.40\pm1.84$&$72.71\pm3.83$&$73.26\pm2.19$&$74.27\pm1.82$&\bm{$74.98\pm2.26$}\\	
							\textbf{Mean}	
							&$64.36\pm2.69$&$65.78\pm2.16$&$73.09\pm1.93$&$79.21\pm3.11$&$80.49\pm2.26$&$80.27\pm2.13$&\bm{$81.66\pm2.35$}\\	
							\textbf{\uppercase\expandafter{\romannumeral2}}
							&$66.17\pm1.56$&$67.55\pm1.41$&\bm{$81.31\pm2.02$}&$78.24\pm3.63$&$80.08\pm3.42$&$79.62\pm3.36$&$81.27\pm3.42$\\					
							\hline
							\hline
							
							\multicolumn{8}{c}{\textbf{Train-10}}\\
							\hline
							\textbf{FAC}  
							&$70.34\pm2.22$&$73.81\pm1.94$&$79.13\pm1.66$&$88.41\pm1.17$&$89.79\pm1.48$&$89.78\pm1.52$&\bm{$90.93\pm1.48$}\\
							\textbf{FOU}
							&$75.05\pm1.67$&$72.88\pm3.64$&$76.15\pm1.95$&$81.47\pm1.42$&$83.19\pm0.85$&$82.47\pm1.17$&\bm{$84.54\pm0.88$}\\
							\textbf{PIX}	
							&$75.95\pm1.03$&$76.71\pm1.28$&$75.75\pm1.03$&$74.40\pm3.92$&$75.59\pm0.59$&$76.00\pm0.64$&\bm{$76.79\pm0.84$}\\	
							\textbf{Mean}	
							&$73.78\pm1.64$&$74.46\pm2.29$&$77.01\pm1.55$&$81.43\pm2.17$&$82.81\pm1.02$&$82.75\pm1.11$&\bm{$84.04\pm1.08$}\\	
							\textbf{\uppercase\expandafter{\romannumeral2}}
							&$76.81\pm2.62$&$78.97\pm2.42$&$84.42\pm2.34$&$81.52\pm1.39$&$83.07\pm0.84$&$82.34\pm1.13$&\bm{$84.45\pm0.96$}\\
							\bottomrule[2pt]
						\end{tabular}}
					\end{table*}	
					The maximum classification accuracy under optimal
					feature extraction are shown in Tables \ref{T1}-\ref{T4}, and the best results are highlighted in boldface. ``\textbf{Train-}\bm{$M$}" represent the classification accuracy of $M$ training samples for per class, and ``\textbf{Mean}" represent the average classification accuracy of all views of the low-dimensional features obtained  by \uppercase\expandafter{\romannumeral1}.  ``\textbf{\uppercase\expandafter{\romannumeral2}}" represent the classification accuracy of the low-dimensional features obtained  by \uppercase\expandafter{\romannumeral2}.
					\par 
					On the Yale dataset:\par 
					It can be seen from Table \ref{T1} that our MFEDCH results in the maximum classification accuracy in 9 cases, which is 2.62\% to 18.53\% higher than the comparison methods on average. Only in other 3 cases (\textbf{GS} in \textbf{Train-6, 8} and \textbf{\uppercase\expandafter{\romannumeral2}} in \textbf{Train-8}), GMDCCA induces the highest classification accuracy, which is only 2.27\% higher than MFEDCH on average. \par
					
					On the MF dataset:\par 
					It can be seen from Table \ref{T4} that our MFEDCH results in the maximum classification accuracy in 13 cases, which is 1.24\% to 14.93\% higher than the comparison methods on average. Only in other 2 cases  (\textbf{\uppercase\expandafter{\romannumeral2}} in \textbf{Train-6, 8}), GDMCCA induces the highest classification accuracy, which is only 0.69\% higher than MFEDCH on average. \par
					From the above experimental results, it can be found  that our MFETCH has obvious advantages compared to both traditional multi-view feature extraction methods and traditional deep learning methods. In particular, by comparing the experimental results of MFEDCH and CMC, it can be known that our method has stronger advantages mainly attributed to the role of the feature- and recovery-level contrastive losses.\par 
					
					\bibliography{bibfile}
					\end{document}